# Reward driven discovery of the optimal microstructure representations with invariant variational autoencoders

Boris N. Slautin[1,*], Kamyar Barakati[2], Hiroshi Funakubo[3], Maxim A. Ziatdinov[4], Vladimir V. Shvartsman[1], Doru C. Lupascu[1], Sergei V. Kalinin[2,4,*]

[1] Institute for Materials Science and Center for Nanointegration Duisburg-Essen (CENIDE), University of Duisburg-Essen, Essen, 45141, Germany

[2] Department of Materials Science and Engineering, University of Tennessee, Knoxville, TN 37996, USA

[3] Department of Material Science and Engineering, Tokyo Institute of Technology, Yokohama 226-8502, Japan.

[4] Pacific Northwest National Laboratory, Richland, WA 99354, USA

**Abstract**

Microscopy techniques generate vast amounts of complex image data that in principle can be used to discover simpler, interpretable, and parsimonious forms to reveal the underlying physical structures, such as elementary building blocks in molecular systems or order parameters and phases in crystalline materials. Variational Autoencoders (VAEs) provide a powerful means of constructing such low-dimensional representations, but their performance heavily depends on multiple non-myopic design choices, which are often optimized through trial-and-error and empirical analysis. To enable automated and unbiased optimization of VAE workflows, we investigated reward-based strategies for evaluating latent space representations. Using Piezoresponse Force Microscopy data as a model system, we examined multiple policies and reward functions that can serve as a foundation for automated optimization. Our analysis shows that approximating the latent space with Gaussian Mixture Models (GMM) and Bayesian Gaussian Mixture Models (BGMM) provides a strong basis for constructing reward functions capable of estimating model efficiency and guiding the search for optimal parsimonious representations.

[*] Authors to whom correspondence should be addressed:  boris.slautin@uni-due.de and sergei2@utk.edu

# I. Introduction

Various microscopy techniques typically represent experimental outcomes as 2D or 3D images that map the distribution of target properties across the surface or within the volume of the investigated material. This makes image analysis a central and unifying task for a diverse array of experimental techniques, from optical to electron microscopy and scanning probe methods.[1-3] Extracting meaningful information from an image often involves deconstructing it into a set of fundamental descriptors constituting primitive elements or building blocks that collectively capture the structural and functional content of the visual data.[4-7]

Interestingly, this principle holds not only for microscopy but also for everyday scenes, for instance, a photo of a bustling city street can be broken down into well-defined common elements such as buildings, cars, people, trees, and more. By analyzing the presence, quantity, variability within each class, and their spatial relationships, it becomes possible to extract a wide range of scene attributes spanning from basic ones like noise, pollution, or traffic density to more complex properties such as social activity levels, urban planning patterns, etc.

The identification of elementary descriptors is a fundamental task in microscopy image analysis. In practice, a descriptor often corresponds to a sub-region of the larger microscopy image that encodes a structural element of interest. Depending on the imaging modality and material system, such elements may include atomic arrangements, molecular configurations, ferroelectric domains, grain boundaries, or other microscale features.[8-11] Identifying these descriptors often relies on prior knowledge and assumptions about the system under investigation. For instance, in scanning transmission electron microscopy (STEM), high-contrast peaks are typically interpreted as atomic positions within the lattice.[12-14] In scanning probe microscopy (SPM), convolutional neural networks as well as morphological transformations, for instance Canny edge detection or Sato filtering, can reveal features like domain walls or ferroelastic domains.[6, 15, 16] These forms of prior knowledge guide the selection of key points and help determine the appropriate size and location of image patches used to deconstruct the image into meaningful structural elements.[8] In other cases, particularly when characterizing continuous or disordered structures, such as amorphous materials or metallographic images, defining suitable descriptors becomes significantly more challenging due to the lack of clear periodicity or distinct features. The construction of an appropriate set of descriptors is a pivotal step that directly impacts the overall efficiency of image analysis workflows, especially those that rely on machine learning (ML) to uncover hidden patterns or structure–property relationships.

**II. Parsimonious discovery in image data**

Reducing a microscopy image to a set of descriptors constitutes a parsimonious representation, where complex microstructures are modeled as a superposition of simpler elements, while retaining their intrinsic variability.[17] In image analysis, such variability is often captured through basic affine transformations, including rotation, translation, and scaling. Descriptors, defined as image patches, are then often encoded into a low-dimensional latent space. These compact representations reduce computational cost, enhance interpretability, and facilitate downstream analysis tasks such as clustering, classification, etc. A range of methods can be used for this purpose, from traditional statistical approaches like Principal Component Analysis (PCA)[18] to more advanced machine learning models such as Variational Autoencoders (VAE),[19] Deep Kernel Learning (DKL),[20] and others.[21] These types of workflows have been widely implemented in electron microscopy and scanning probe microscopy (SPM), forming a foundation for both data analysis and automated experimentation.[17, 22-25]

Among these methods, the VAE is particularly promising for dimensionality reduction. The encoder network in a VAE captures nonlinear variability in the data and maps it into a low-dimensional latent space. The generative nature of the VAE enables the creation of out-of-distribution structures by sampling from the latent space, which is an essential capability for many optimization tasks.[26-28] While direct interpretation of VAE latent variables can be challenging, extending the standard VAE architecture with rotational and translational invariance enables the model to isolate these well-defined geometrical factors.[23, 29] This improves the interpretability of the method and encourages the model to capture hidden, physically meaningful variations in the latent space.

The VAE-based image analysis workflow in microscopy can be structured as a multistage process (Figure 1a). In the first stage, a system of key points is defined from which descriptors are constructed. These key points can be selected using either simple grid-based sampling strategies or established feature detection algorithms, such as ORB[30] or SIFT.[31] Physically, the key points can correspond to atomic positions in atomically resolved images,[25, 32] sampled points at grain boundaries,[6] or ferroelectric domain walls.[33, 34] Such a key-point system is specifically tailored to represent the material properties of interest. The next stage involves defining the size of the image patches (descriptor size), which is a central hyperparameter in the VAE workflow. Ideally, each descriptor should capture a single structural element. In the final step, the constructed descriptor system is encoded into a low-dimensional latent space using a selected VAE model. Depending on the chosen invariants, the workflow output are latent variable maps ($z_1, z_2$ ...) extended by the corresponding rotation angles ($\theta$) and/or translation vectors ($t_x, t_y$) when using rotationally and/or translationally invariant VAEs (Figure 1b-d). The VAE latent distribution provides a parsimonious representation of the microscopic data, which can be further utilized in automated experimental workflows, for example, as a basis for selecting measurement locations.[25, 35, 36]

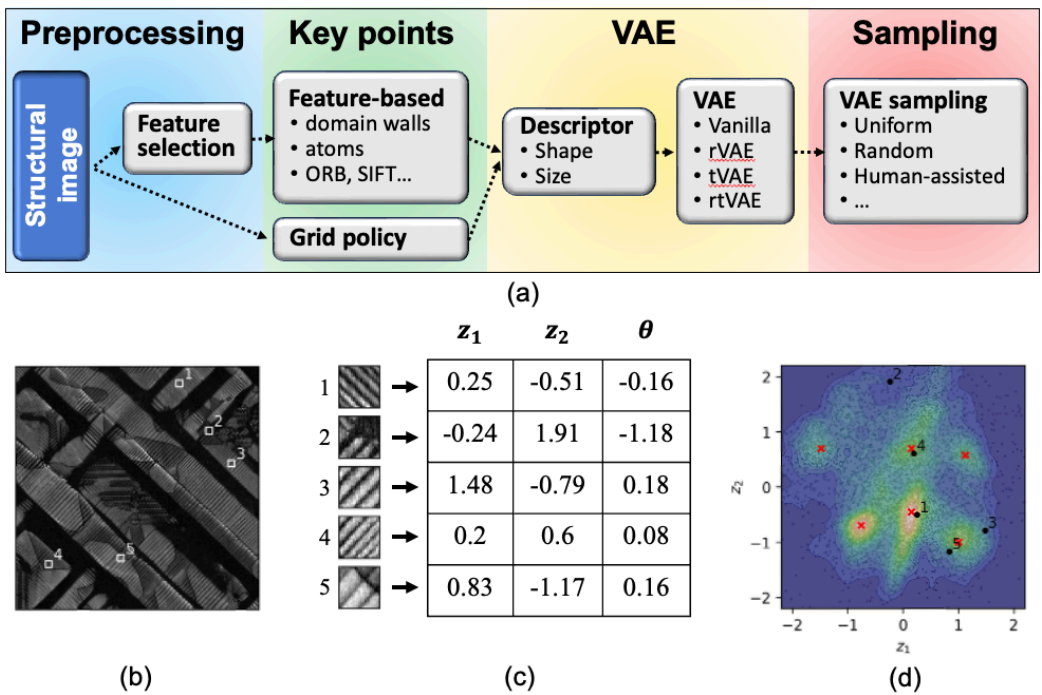

**Figure 1.** (a) Schematic representation of the multi-stage VAE-based workflow for achieving an optimally parsimonious representation of microscopy data. (b–c) Schematic of VAE encoding of ferroelectric domain structures in a PbTiO₃ film: (b) PFM amplitude image, (c) r-VAE latent representations of random image patches capturing local domain arrangements, and (d) the corresponding r-VAE latent space. Distinct localized clusters are observed in the latent space, with density maxima marked by red crosses. These clusters correspond to discovered microstructures, while the manifolds connecting them represent intermediate configurations.

Even in relatively simple VAE workflows, the outcomes are strongly influenced by choices made in the early stages (non-myopic workflow). For instance, construction of the key-point system, even in case when it can be guided by the underlying physics or specific area of interest, involves tuning of multiple hyperparameters which significantly affect the downstream outcomes.

The choice of VAE model balances the benefits of disentangling geometric factors against the added computational complexity. The combination of these factors and possible complexity of subsequent workflow configurations (for automated experimentation, advanced analysis, etc.) can make the resulting analysis suboptimal and prone to user bias. Additionally, in practice, the outcomes available for interpretation emerge only at the end of the workflow, at which point they already reflect the compounded influence of multiple prior decisions. Hence, we need strategies to construct parsimonious representations in a manner that is unbiased and requires minimal human intervention.

Here, we investigate various reward-driven strategies for optimizing VAE-based image analysis workflows to achieve parsimonious representations of microscopy data. The proposed approaches are demonstrated using piezoresponse force microscopy (PFM) data.

**III. VAE\*: Reward based VAE analysis**

Automated optimization of VAE workflows requires defining both the parameter space and the reward function that guides the search. The core idea of a reward-driven workflow is to express the desired outcome as a quantifiable reward, which directs the optimization process.[37-40] Rather than relying on predefined algorithmic steps or heuristic choices, candidate solutions are evaluated by how well they maximize this reward. In our case, optimization is formulated as a combinatorial exploration of the entire workflow structure, comprising a sequence of discrete operations, where the hyperparameters of individual operations are jointly optimized within a unified search space.

The central challenge of this work is to define rewards that can effectively guide the optimization process toward achieving an optimal parsimonious representation. In turn, constructing such rewards requires identifying the underlying principles that distinguish efficient representations from suboptimal ones. Hence, for the VAE-based workflow, it is important to define fundamental criteria for latent-space representations. Considering the initial microscopy image as a superposition of elementary structural blocks, we can formulate the following requirements for a parsimonious VAE representation:

1. the latent representation should retain as much relevant information about the system as possible;
2. elementary structural elements should remain distinguishable in the reduced-dimensional space;

3. simpler latent representations are preferred over unnecessarily complex ones.

Optimization of the VAE workflow to balance these criteria can be achieved either through a single reward function, that combines all criteria, or via multi-objective optimization, where each reward corresponds to an individual criterion. The latter approach enables exploration of the Pareto front to identify optimal trade-offs between objectives.[41-43]

Below, we discuss several available functions, ranging from simple to more complex, that can be used for reward construction. While our primary focus is on developing a single reward approach, we also consider functions that may not fully satisfy all the outlined criteria but can serve as individual rewards within a multi-objective optimization framework.

### III.1. Reconstruction loss

An employing of the VAE reconstruction loss as a reward function is one of the most intuitive choices. The final reconstruction loss measures difference between the input and output of VAE model, reflecting how effectively the model can compress high-dimensional data into a lower-dimensional latent representation, in other words, how much information the latent space is able to capture.

The reconstruction loss tends to be sensitive to key hyperparameters in the workflow, particularly the architecture of the encoder/decoder networks and the size of the input descriptors. When using negative log-likelihood as the reconstruction loss, the corresponding reward function can be expressed as follows:

$$Reward(Z|\boldsymbol{p}) = -\mathcal{L}_{rec} = \frac{1}{N}\mathbb{E}_{q_\phi(Z|X)}[\sum_{i=1}^{N} \log p_\theta(x_i|z)] = \frac{1}{N}\mathbb{E}_{q_\phi(Z|X)}[\log p_\theta(x|z)], \quad (1)$$

where $N$ is the number of the elements (pixels) in descriptor, $\mathbb{E}_{q_\phi(Z|X)}$ the expectation over the latent space, and $p_\theta(x_i|z)$ the decoder likelihood.

The reconstruction loss only estimates the quality of the encoding (1st criterion) making it suitable only for multi objective workflows.

### III.2. Latent distribution variance

Latent distribution variance is a parameter for detecting latent space collapse, when all input descriptors project into a single point in the latent space. While the distribution total variance, defined as the sum of variances across all latent dimensions $Var(Z) = \sum Var(z_i)$, can indicate general collapse, it fails to reveal the collapse of individual latent variables, if others remain non-zero. To address this, a surrogate variance-based reward can be established as the product of individual variances across all latent dimensions:

$$Reward_{var}(Z|\boldsymbol{p}) = \prod_i \text{Var}(z_i) \qquad (2)$$

Collapse or low variance of individual latent variables may indicate excess dimensionality in the latent space and can serve as a criterion for tuning this hyperparameter. At the same time, optimizing the overall workflow variance alone does not provide a comprehensive measure of the desired criteria and therefore rarely serves as an effective reward function.

**III.3. Gaussian Mixture Model**

Regularization of the latent space is a central feature of VAEs that enables their generative capabilities. Typically, the latent space is regularized using a standard normal prior $N(0, I)$. During training, an additional Kullback–Leibler (KL) divergence term is included in the loss function alongside the reconstruction loss. This KL term encourages the learned latent distribution to align with the prior, effectively shaping the structure of the latent space and promoting smooth, continuous sampling for generation.

While minimizing the KL term is intended to encourage the latent distribution to align with the standard normal prior $N(0, I)$, in practice, the learned latent representations often exhibit clustering behavior. Importantly, VAE does not directly promote clustering and any clustering observed is mostly a consequence of the internal organization of the encoded system. These clusters reflect the presence of distinct types of descriptors (e.g., elementary microstructures) within the input data. Since each descriptor is expected to represent an individual structural unit, a latent representation that naturally forms distinct clusters can be regarded as an effective parsimonious description of the system and satisfies the criteria for an optimal parsimonious representation. Assuming that each cluster can be approximated by a multivariate Gaussian distribution, a Gaussian Mixture Model (GMM) provides a natural framework for analyzing the latent space. The Gaussian-like shape of distinct clusters in the latent space stems from the VAE architecture: each input is encoded as a multivariate normal distribution, and the aggregation of such encodings for similar inputs results in clusters that reflect mixtures of these distributions. Nevertheless, latent space clusters are not constrained to adopt a strictly single-component

Gaussian form, which makes this assumption valid only as an approximation. The Gaussian-like shape of each cluster reflects a continuous representation of the variability within individual structural elements, analogous to how the Gaussian prior on the entire latent space encourages overall continuity in the representation.

Considering the above, the reward function can be formulated using a quality metric that assesses how well a GMM approximates the latent representation. The Bayesian Information Criterion (BIC)[44] and Akaike Information Criterion (AIC),[45] which combine a model fit with a penalty for excessive complexity, are of particular interest for this purpose.

$$Reward_{BIC}(Z|\boldsymbol{\theta}) = BIC(Z|\boldsymbol{\theta}) = -2\log(\mathcal{L}) + k\log(n) \quad (3)$$

$$Reward_{AIC}(Z|\boldsymbol{\theta}) = AIC(Z|\boldsymbol{\theta}) = -2\log(\mathcal{L}) + 2k \quad (4)$$

were $Z$ is VAE latent distribution, $\boldsymbol{\theta}$ represents the tunable parameters, $\mathcal{L}$ is the likelihood, $k$ the number of free parameters in the GMM model, and $n$ the number of data points in $Z$. While both criteria balance the model fit and the complexity, BIC applies a stronger penalty for complexity, making it less prone to overfitting. In the following analysis, we focus on BIC, although the discussion is largely applicable to AIC as well.

There are several limitations to using BIC as a reward function. In an ideal scenario where the VAE latent space perfectly follows the standard normal prior *N(0,I)*, a single-component GMM provides the best fit with minimal complexity, resulting in the lowest BIC value. A multi-cluster latent distribution requires multiple Gaussian components to achieve a good fit, which increases the number of free parameters and typically leads to a higher BIC. This can bias the optimization toward favoring a single-component latent representation, potentially discarding informative multi-modal structures. Thus, constructing a BIC-based reward function requires additional regularization to avoid favoring excessively simplistic latent representations.

The second limitation arises from the dependence of both the likelihood and the complexity penalty in BIC on the number of samples. This dependence makes it impossible to directly compare BIC values across distributions with different sample sizes. One potential solution is to normalize the BIC score by the number of samples ($BIC/N$). Although this normalization disrupts the original balance between the reconstruction and complexity terms, it can be a reasonable approximation when comparing large distributions with only moderate differences in sample size.

Alternatively, BIC scores can be compared using representative subsets with equal sample sizes, ensuring a fairer evaluation.

A final limitation of the BIC and AIC scores lies in the absence of any penalty for overlapping clusters. In representing the system as a set of elementary structural elements, the ideal case is to fit each element type with a single component. However, as discussed above, clusters in the latent space are better described as mixtures of multivariate normals. Hence, in some cases, BIC and AIC may prioritize approximating a single cluster with multiple GMM components over fitting it with a single component. While such a multicomponent representation can capture the variability within each type of structural element, it may also obscure the intended one-to-one correspondence between clusters and structural elements.

The last limitation can be partially addressed by employing the Integrated Completed Likelihood (ICL),[46] which extends the BIC score by incorporating the entropy of the cluster assignments ($H$). As a result, ICL favors GMM models with well-separated clusters, reducing the tendency to approximate a single latent-space cluster with multiple components. The ICL score can be expressed as:

$$Reward_{ICL}(Z|\boldsymbol{\theta}) = BIC(Z|\boldsymbol{\theta}) - H(Z|\boldsymbol{\theta}) = BIC(Z|\boldsymbol{\theta}) - \sum_{n=1}^{N}\sum_{k=1}^{K} \tau_{ik}\log(\tau_{ik}), \quad (5)$$

where $\tau_{nk}$ is the posterior probability of assigning $z_i$ to the GMM component $k$.

All three GMM-based metrics – BIC, AIC, and ICL – formally favor distributions that satisfy the requirements for an optimal parsimonious representation. Therefore, they can serve as reward functions

**III.4. Bayesian Gaussian Mixture Model: Evidence Lower Bound**

The Bayesian Gaussian Mixture Model (BGMM) extends the standard GMM by introducing Bayesian inference, enabling direct incorporation of priors over its parameters. Unlike the standard GMM, BGMM produces probabilistic scores that quantify how well a given distribution aligns with the imposed priors. This allows the explicit formulation of criteria for an optimal latent space and mitigates the tendency to favor a trivial $N(0, I)$ latent distribution.

The Evidence Lower Bound (ELBO) from fitting the BGMM to a given distribution serves as the foundation for constructing the reward function. The ELBO combines a likelihood term, which evaluates the quality of the fit, and a KL divergence term, which measures the alignment of the model with the specified priors:

$$Reward_{ELBO}(Z, \boldsymbol{\theta}) = ELBO = \mathbb{E}_{q_\phi(\bar{z}|z)}[\log p_\theta(z|\bar{z})] - D_{KL}(q_\phi(\bar{z}|z)||p(\bar{z})) \qquad (6)$$

where $z$ represents the input data, and $\bar{z}$ is the latent representation under the BGMM. The function $q_\phi(\bar{z}|z)$ is the approximate posterior distribution of $\bar{z}$ conditioned on $z$, parameterized by $\phi$, $p_\theta(z|\bar{z})$ is the likelihood of $z$ given $\bar{z}$, and $p(\bar{z})$ is the prior distribution over the BGMM components. The term $D_{KL}(q_\phi(\bar{z}|z)||p(\bar{z}))$ refers KL divergence between the conditional distribution $q_\phi(\bar{z} \mid z)$ and the prior $p(\bar{z})$. Unlike VAE model training, the prior for the BGMM may reflect multi-clustering distributions representing the optimal organization of the latent space. When comparing ELBO values across latent distributions with different sample sizes, it is necessary to normalize by the number of samples or evaluate the ELBO on representative subsets of equal size.

The distinctive feature of BGMM is its intrinsic ability to optimize the number of components in the distribution through the Dirichlet prior over the component mixing weights, whereas in a standard GMM the number of components must be specified explicitly. The concentration parameter of the Dirichlet prior controls sparsity: selecting a small concentration suppresses redundant clusters and encourages parsimony.

The BGMM ELBO serves as a highly flexible reward function, allowing simultaneous incorporation and balancing of multiple requirements: the quality of the latent encoding through the log-likelihood term, and the complexity and distinguishability of structural elements through the priors. However, like the BIC and AIC scores, the ELBO does not include a penalty for overlapping clusters. A potential solution is to construct a surrogate reward function that combines the ELBO with an additional penalty, such as the entropy of the cluster assignments.

Beyond the rewards discussed above, we also explored complementary approaches such as topological analysis. However, these rewards did not recommend themselves as efficient criteria for scoring the VAE-based workflow. Jupyter notebooks, including our experiments and a flexible VAE-workflow constructor for incorporating and testing user-defined reward functions, are provided (see Data Availability).

**IV. Ferroelectric domain analysis**

We applied the proposed reward functions for VAE workflow optimization using local domain distributions visualized via Piezoresponse Force Microscopy (PFM) in a PbTiO$_3$ thin film (Figure S1a). The analyzed PFM scan of the PbTiO$_3$ crystal represents a diverse domain structure consisting of *c*-domains and *a*-domains, characterized by out-of-plane and in-plane polarization directions, respectively. The dense domain structure and variability in local domain arrangements within the film make the identification of an efficient, parsimonious representation a challenging task.

VAE models with different invariances but containing the same structure have been trained for different descriptor sizes ($ds$). In case of the square descriptors (image patches), the descriptor size is defined by the window size $ds = ws^2$. Importantly, the reconstruction term of the VAE ELBO loss scales with the descriptor size, whereas the KL divergence generally remains constant. Therefore, to isolate the direct influence of the descriptor size from second-order effects, such as changes in the ratio between the loss terms during training, we scale the KL term proportionally to the descriptor size. The $ds$ and choice of the invariants – vanilla VAE, rotationally invariant (r-VAE), or rotationally and translationally invariant (t-VAE) – define an optimization space. The selection of these parameters for optimization is motivated by the fact that they are typically tuned empirically, whereas the optimization of many other parameters may be guided by underlying physical principles. For instance, key points for constructing the descriptor system were selected on domain walls, as these are the main objects of interest in many ferroelectric applications (Figure S1b). All models possess the same internal structure and training hyperparameters; a detailed description of the model architecture and training process is provided in the Materials and Methods section.

Changing the descriptor size leads to variations in the latent distributions (Figure 2). In many cases, these distributions appear as distinct, high-density clusters surrounded by more diffuse point clouds. The dense clusters are likely to represent specific types of domain walls, whereas the surrounding clouds may correspond either to underrepresented microstructures or to structures composed of multiple elementary building blocks. For the smallest window size, $ws = 3$, we did not observe the formation of distinct clusters in the vanilla VAE, while only two clusters appeared in the r-VAE and t-VAE, highlighting the limited variability captured at this window size. A similar trend of decreasing clusterization, though less pronounced, was observed for larger window sizes ($ws > 11$), particularly in the r-VAE and t-VAE. We attribute this to the potentially excessive size of the descriptors, which leads to mixing of different structural elements within a single image patch and thereby reduces their separability. The presence of distinct cluster structures in the vanilla VAE for large $ws$ can be explained by the strong influence of domain wall orientation, which becomes the primary factor shaping cluster formation in the latent space. This geometrical factor is isolated into a single variable and is not included in the latent spaces of the r-VAE and t-VAE.

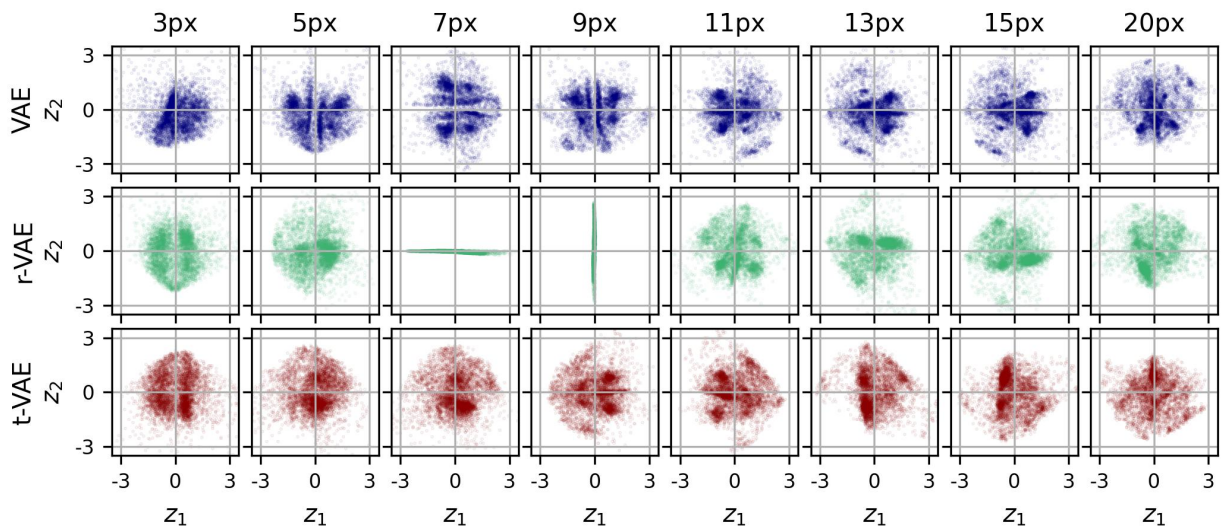

**Figure 2.** VAE latent distributions vs. model invariances and window size.

Collapses of the latent space along one variable were observed for $ws = 7$ and $ws = 9$ in the r-VAE. These collapses can be attributed to inefficient encoding or to the excessive dimensionality of the latent space. Apart from the collapsed distributions, the latent spaces of the t-VAE and r-VAE exhibit well-defined similarities across different window sizes. This behavior may be explained by the selection of key points along the domain wall, which reduces the influence of translational invariance.

**IV.1. Reconstruction loss**

For the reconstruction loss, we observe a steady increase with window size (Figure 3a). Enlarging the image patch increases the amount of information that must be encoded and, consequently, the complexity of the encoding process. Thus, the observed growth is expected. For most of the window sizes the reconstruction loss of t-VAE and r-VAE lies below those for vanilla VAE. The isolation of the geometrical factors effectively increases the dimensionality of the latent space making it possible to encode more variability factors. Overall, the observed dependency supports our assumption that, while reconstruction loss is rarely suitable for optimizing descriptor size, it can serve as a pivotal metric for tuning other parameters, such as the VAE architecture and the choice of invariances, by balancing the improvement in reconstruction performance against the associated increase in computational complexity.

**IV.2. Variance**

The total variances of the VAE latent distributions, excluding collapsed cases, show only a weak dependence on window size (Figure 3b), fluctuating around the expected value of 2 for a 2D latent space. We expect that collapses along a single dimension should result in a variance drop of about 1 per collapsed variable, which can be used to estimate the effective dimensionality of the latent space. For instance, the collapse of a single latent variable in the r-VAE with $ws = 7$ or $ws = 9$ reduces the total variance to approximately 1. The variance-based reward, defined as the product of the variances along individual latent variables (Equation 2 and Figure 3c), drops sharply towards zero when a collapse of any latent variable occurs. Therefore, it is useful primarily for detecting collapses in general rather than for providing a detailed characterization of the latent space.

While variance-based scores cannot serve as an independent reward function for window size optimization, they can be used to optimize the dimensionality of the VAE latent space and to filter collapsed distributions. This step is pivotal for employing GMM- and BGMM-based rewards, as demonstrated below.

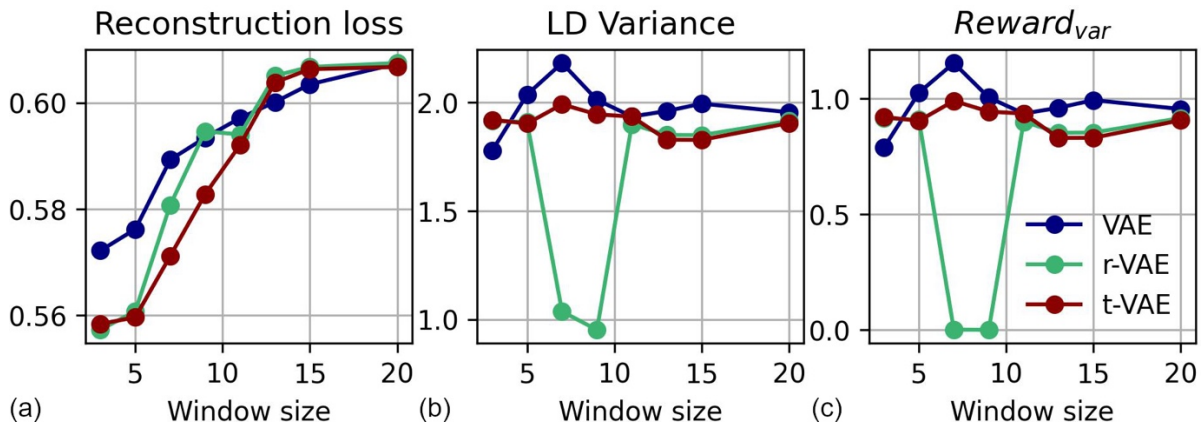

**Figure 3.** Dependence of (a) VAE reconstruction loss, (b) latent distribution variance, and (c) variance-based reward function on window size for different VAE invariances.

### IV.3. GMM

To examine the applicability of GMM-based scores, we applied GMM fitting to all latent distributions. As described above, we assume that each GMM component corresponds to a distinct microstructural type. However, the presence of low-density point clouds in the latent distributions, corresponding either to underrepresented microstructures or to patches combining several elementary microstructures, often leads to components with large variances, effectively serving as a nonlinear background and hampering further analysis. To eliminate this effect, we applied density-threshold filtering: points from regions with density below 95% of the maximum density in each distribution were removed, and the filtered distributions were then used for GMM training (Figure S2). The exact number of GMM components is a hyperparameter that must be defined before fitting. To determine the optimal number of components, we performed a grid search over the range of components from 2 to 10, selecting the best value for each window size based on the chosen reward function. The resulting optimal GMM models were then compared across different window sizes to identify the global optimum. Since the number of points in distributions decreases with increasing window size, to eliminate dataset-length effects we performed GMM fitting on representative subsets, each equal in size to the smallest filtered latent distribution.

First, we employ the GMM log-likelihood as a reward function. The log-likelihood evaluates only the quality of the GMM fit, while ignoring model complexity and providing no penalty for overlapping components. The absence of model complexity accounting leads to favoring GMMs with a larger number of components, typically around 8–9 for most invariance types and window sizes, which is close to the upper bound of the examined range (Figure 4b).

Interestingly, the latent distributions of the r-VAE with $ws = 7$ and $ws = 9$, where collapse along one of the latent variables occurred, received the highest scores. The same effect was observed for all examined GMM-based rewards (Figure 4a). The dense latent distributions arising from collapse, where each point is effectively sampled from its own narrow normal distribution, possesses a simple shape, leading to their prioritization by reward functions. This highlights the limited applicability of GMM-based rewards in this context and necessitates the use of additional metrics, such as the variance-based rewards discussed above, to identify latent space collapses. It is important to underscore that collapse may arise not only from ineffective encoding but also as an indication of excess latent dimensionality. However, this lies beyond the scope of the present window size optimization. For the purposes of this article, we excluded collapsed latent distributions from consideration, interpreting them as artifacts of ineffective encoding rather than as intrinsic physical effects.

The t-VAE model with the largest window size in the examined range achieved the highest score based on the log-likelihood function. Excluding collapsed cases, the r-VAE and t-VAE show very similar scores for small window sizes, which can be attributed to the specific of the key-point selection method at the domain walls. However, for $ws > 13$, the difference becomes more pronounced, highlighting the increased importance of translational invariance in this range.

Employing BIC as a reward function results in a slight decrease in the number of components, while the shape of the BIC dependence on window size closely mirrors that of the log-likelihood. The reduction in components arises from the complexity penalty, whereas the similarity of the curves reflects the pivotal role of log-likelihood in the BIC reward. The BIC criterion also prioritizes the t-VAE with the largest window size ($ws = 20$). Introducing an entropy term in the ICL reward function leads to a dramatic decrease in the number of components, as the penalty is applied to overlapping clusters. For many r-VAE and t-VAE models, especially at small $ws$, the number of clusters dropped to two, that is the minimum of the examined range. In contrast, vanilla VAE models were noticeably less prone to such drastic reductions, which can be attributed to geometrical factors introducing additional variability. Ultimately, employing ICL shifts the optimal window size from 20 px to 13 px.

Additionally, we examined ICL-based functions with artificially elevated importance of the entropy term by dividing the BIC term by 2 ($ICL_{1/2}$) and by 4 ($ICL_{1/4}$). In our case, this did not lead to a significant change in the prioritization of models; however, we believe that reward functions with a reduced emphasis on fit quality may be suitable for favoring a more distinguishable cluster structure in latent space over strictly Gaussian-shaped clusters.

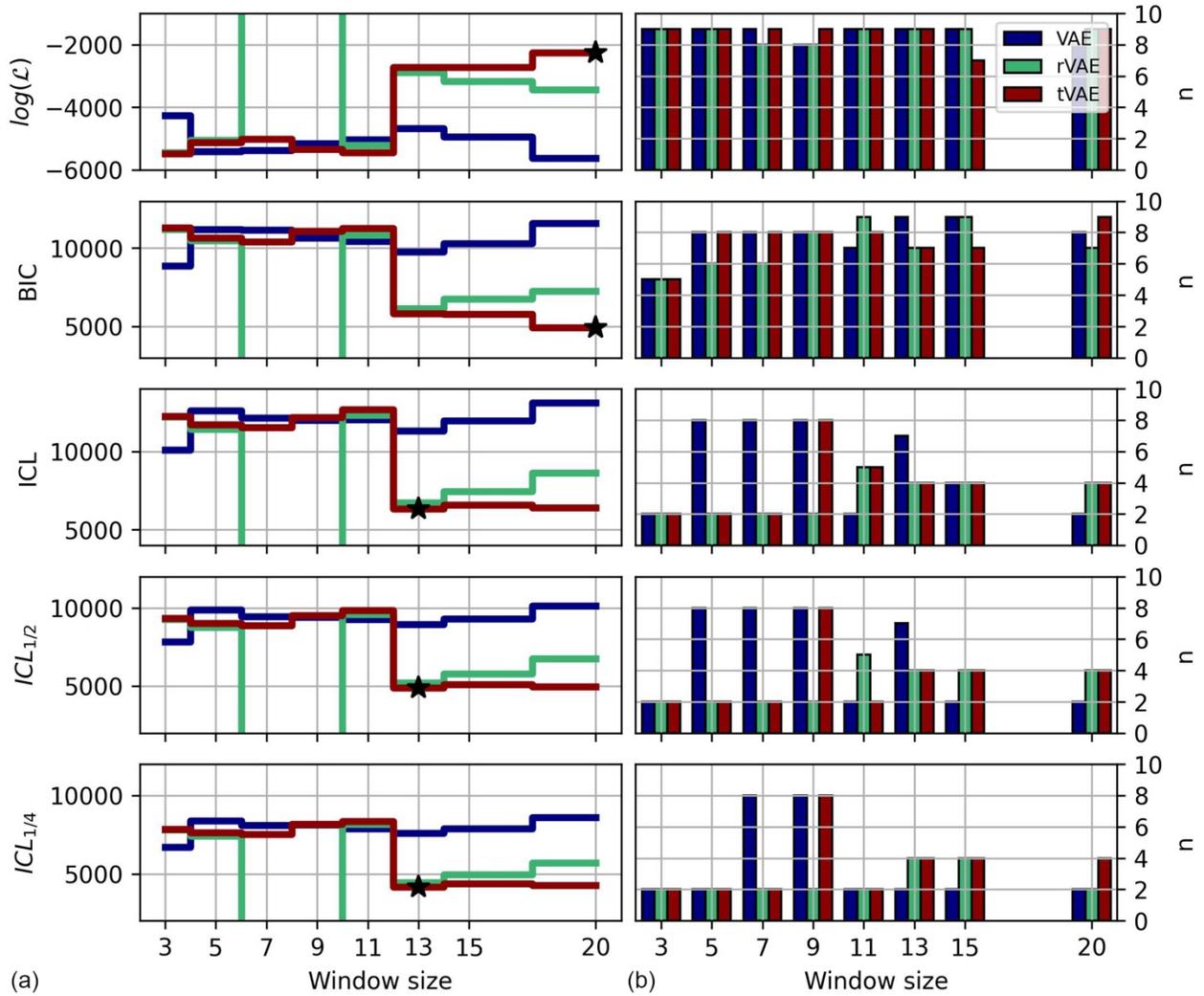

**Figure 4.** GMM-based reward functions. (a) Different reward metrics and (b) the optimal number of components (*n*) in GMM models as a function of window size. The optimal window sizes and invariances are highlighted with a star for each reward function.

The latent distributions of the t-VAE models with $ws = 13$ and $ws = 20$ were identified as optimal by the ICL-based metrics and by BIC (and log-likelihood), respectively. To evaluate the efficiency of GMM fitting for selecting an optimal parsimonious representation and identifying the core microstructures in the analyzed PFM scan, we analyzed the distribution of GMM components in the latent space and their corresponding reflections in real space (Figure 5). As examples, we selected the GMM model chosen by the $BIC$ reward, approximating the t-VAE with $ws = 20$, and the model chosen by $ICL_{1/4}$, approximating the t-VAE with $ws = 13$. It is important to remember, that all models were trained on the filtered latent distributions.

The GMM trained on the latent space with $ws = 20$ exhibits a diverse structure, with Gaussian components of varying sizes and covariance shapes. Considering $2\sigma$ as the effective boundary of each component cluster, we observed noticeable overlap between some of them (Figure 5a). The 25% of points located nearest to the cluster centers have been defined as *core points* and we expect that they represent the clearest microstructural types present in the scan. The analysis of their mapping to real space shows that the core points within each cluster correspond to similar microstructures (e.g., a–c domain walls between large domains, a–c domain walls between needle-like domains, c–c domain walls, etc.). However, the criteria for distinguishing some clusters remain unclear. For instance, the core points from the clusters marked in cyan, yellow, and violet all correspond to needle-like a–c domain walls (Figure 5b). While subtle differences between them may exist, representing them as distinct microstructural types appears excessive and deviates from the principle of parsimony in representation. This separation can arise from the absence of a penalty on overlapping components, which in this case leads to capturing intra-microstructure variability rather than combining an entire microstructure type within a single component.

The GMM model trained on the latent space with $ws = 13$ demonstrated a more uniform component structure, with fewer but larger components (Figure 5c). Effectively, the distribution was approximated by only four components. Analysis of the core point mapping to real space shows that each component corresponds to a distinct microstructural type. Points from the clusters marked in cyan and yellow map to the a–c domains of the needle-like a-domain corrugation, with the difference between them likely reflecting opposite polarization directions of the c-domain. Red points correspond to a–c domain walls lie in the border large strip-like a-domains, while violet points are mostly aggregated in the vicinity of c–c domain walls (Figure 5d).

In our view, the VAE representation selected by the ICL-based metrics better reflects the idea of parsimony. We therefore consider ICL-based rewards, which combine the estimation of Gaussian fit quality in the latent distribution (log-likelihood term), model simplicity (BIC complexity term), and cluster separability (entropy term), as the most promising for application as reward functions.

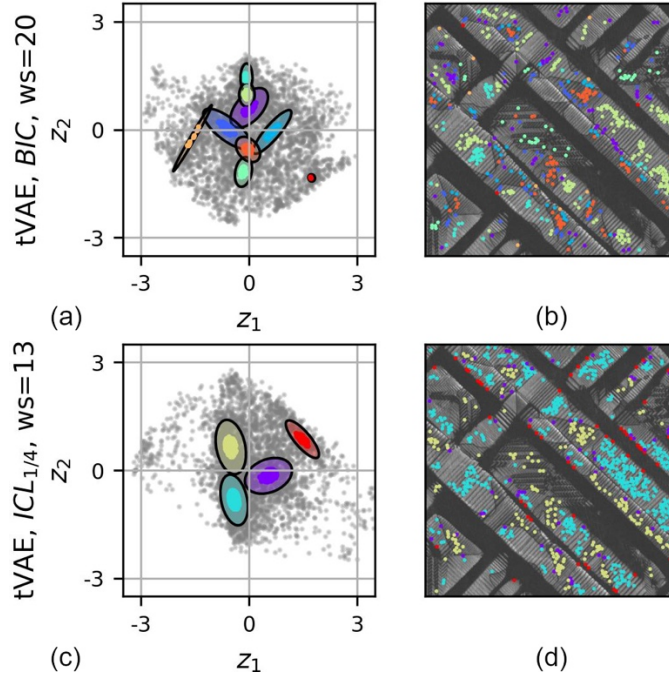

**Figure 5.** t-VAE latent distributions selected with BIC (a) and $ICL_{1/4}$ (c). Ellipses mark GMM components at the $2\sigma$ level, and colored points indicate core samples (25% closest to the cluster center). Panels (b,d) display PFM patch centers corresponding to the BIC- and ICL-based selections.

**IV.4. BGMM**

The Bayesian GMM reward offers multiple opportunities to incorporate desired parameters as priors during fitting. These include, for example, specifying the region of the mean cluster distribution, the expected cluster size, and the covariance structure. By adjusting the concentration parameter of the Dirichlet prior on the component weights, we can also control the sparsity of the model, thereby promoting the removal of excessive components. However, this flexibility comes at the cost of the existence of multiple priors that must be defined before fitting. It should be noted that the introduction of such priors, which ultimately shape the fitting results and thus the reward score of the distribution, somewhat contradicts the core idea of automated and independent discovery of optimal parsimonious representations. Nevertheless, since some priors can be standardized and made system-independent across all VAE representations, we believe that the BGMM model remains a promising basis for reward functions.

The BGMM models were trained for each window size with the examined range allowing a maximum of 20 components. However, the effective number of components was defined as the number of components whose weights exceed 20% of the maximum component weight in the model. The following probabilistic priors were specified: the Dirichlet concentration for the component weights was set to 0.01 to promote sparsity of the model. The means of the cluster centers were drawn from a $Normal(-5,5)$ distribution, effectively covering the whole latent space; the LKJ distribution with concentration parameter 20 was used for the covariance matrices, promoting nearly spherical cluster shapes; and the component variances were sampled from a $HalfNormal(0.05)$ distribution, which limits the cluster width and thereby hinders adaptation to background noise. Overall, the abovementioned priors should promote approximation of the latent distribution by the set of the medium size spherical components which reflects the desired shape of the latent distribution.

The Evidence Lower Bound (ELBO) was used for BGMM fitting. By combining the expected log-likelihood with the Kullback–Leibler (KL) divergence to the priors, the ELBO effectively guides optimization while incorporating prior information. However, the standard ELBO does not include any penalty for component overlap, which proved to be a pivotal criterion in training standard GMM models. To address this limitation, we introduced an additional Bhattacharyya–coefficient–based overlap penalty, making the effective loss function $L = -ELBO + \lambda \cdot Overlap$, which discourages excessive overlap between components. The $\lambda$ coefficient was set to 0.1.

The number of principal components for the chosen weight threshold generally varied between 8 and 12, which is close to the results obtained with BIC-based GMM rewards (Figure 6a). Using the ELBO as a reward function, we found that it clearly prioritizes the latent distribution of the t-VAE model with $ws = 13$ (Figure 6b). This outcome is consistent with the ICL-based selections. While ELBO does not directly estimate the degree of overlap between GMM components, we calculated the classification entropy for each model (Figure 6c). Interestingly, the average overlap was lower for GMMs applied to the vanilla VAE latent distributions, even though these models received lower scores from both GMM- and BGMM-based rewards. To account for overlap when selecting the optimal distribution and, therefore, the optimal invariances and window size, we propose a surrogate reward that combines classification entropy ($H$) with ELBO. This combined reward incorporates cluster separability in addition to fit quality and should provide a more consistent criterion for selecting latent distributions. While we used a simple summation, the relative impact of $H$ can be tuned by introducing a corresponding weight. The combined $-ELBO + H$ reward favors a window size of $ws = 13$ and yields almost equal scores for all three invariances, with a slight advantage for the r-VAE model.

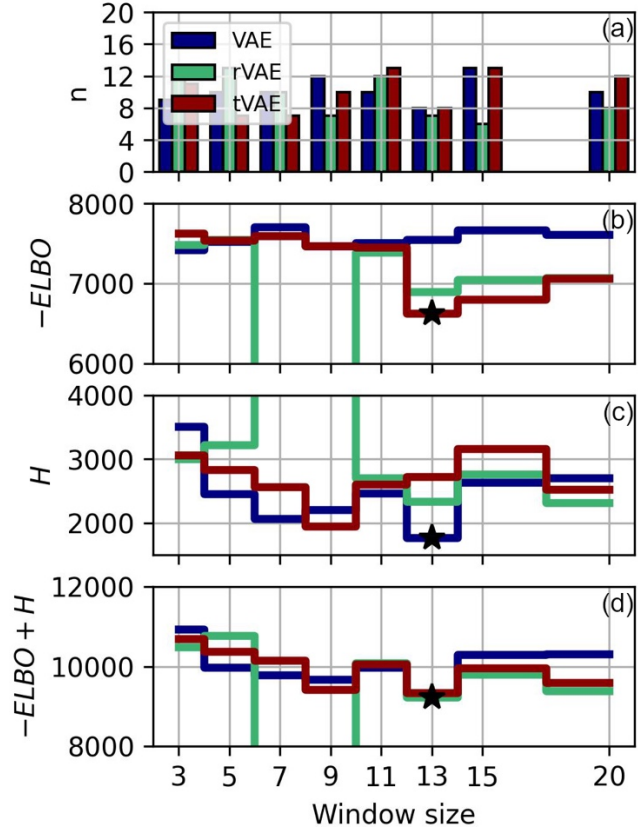

**Figure 6.** BGMM-based reward functions. (a) Number of principal components; (b) $-ELBO$; (c) $H$; and (d) $-ELBO + H$ metrics as functions of window size. The optimal window sizes and invariances are highlighted with stars for each reward function.

The analysis of GMM components in the latent space distribution for the r-VAE with $ws = 13$, favored by the combined reward, shows several partially overlapping clusters of comparable size (Figure 7a). Overall, the individual clusters are larger than those obtained with standard GMM models, and their size variance is much lower, which can be attributed to the chosen priors. Another contributing factor to the larger component sizes is that BGMM fitting was applied to the entire distribution rather than to density-threshold–filtered data. Although partial overlap is evident in the $2\sigma$ ellipses, the core points remain well separated.

Reflection of the core points into real space shows that the model segmented most microstructural types in the latent space with good accuracy (Figure 7b). The core points marked in light green correspond to the domain walls of needle-like a-domains; like the ICL reward, red points lie on the domain walls of striped a-domains, and violet points are mostly associated with c–c domain walls. Interestingly, the a-c domain walls of the herringbone domain pattern are isolated within a single cluster marked in orange. The attribution of the blue points is more challenging, but in many cases these clusters appear near grain boundaries, where visible change of the domain structure is observed. No consistent trend was observed for the cyan cluster, whose points are mostly accumulated near c–c and a–c domain walls of the striped domains; therefore, this cluster can be identified as excessive.

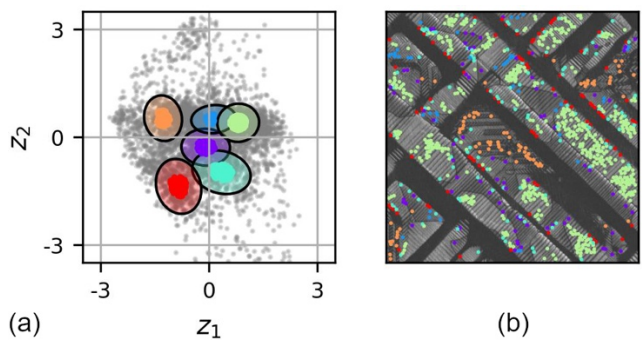

**Figure 7.** (a) r-VAE latent distribution selected with $-ELBO + H$. Ellipses mark GMM components at the 2σ level, and colored points indicate core samples (25% closest to the cluster center). (b) PFM image with patch centers corresponding to the core points.

The combined $-ELBO + H$ BGMM reward highlighted the same window size as the GMM-based *ICL* reward. Since the difference between the t-VAE and r-VAE latent distributions is minimal, we may conclude that a high level of consistency was also observed in the invariance selection. Approximating the latent distribution with Gaussian mixtures through BGMM yielded reasonable results, but with a higher level of excess. Improved hyperparameter tuning of the BGMM should allow for fine-tuning and enhanced performance of the BGMM-based reward.

### V. Summary

Summarizing, we examined various approaches for identifying the optimal VAE-based parsimonious representation of microscopy data and formulated general principles and criteria for selecting such representations. We proposed reward-based strategies grounded in latent space analysis and evaluated multiple reward functions that can enable unbiased and automated VAE hyperparameter optimization. Based on our analysis, approximating the latent space with a mixture of Gaussian functions, each representing a distinct microstructural type, emerges as the most powerful approach for identifying optimal parsimonious representations.

Using Piezoresponse Force Microscopy data, we evaluated GMM- and BGMM-based models for approximating the latent space and for constructing reward scores. Among the tested approaches, the ICL GMM reward and the BGMM reward combining ELBO with an additional component-overlap penalty showed the most promise in identifying and disentangling optimal parsimonious representations. While the efficiency of the proposed framework ultimately depends on the complexity of the system and the diversity of the captured microstructural patterns, we believe this work provides a foundation for further advances in automated and unbiased optimization of VAE-based microscopy workflows.

**Materials and Methods**

The amplitude channel of a PFM scan, measured with an MFP-3D microscope (Oxford Instruments) in Dual Amplitude Resonance Tracking (DART) mode on a thin PTO film, was used to demonstrate the VAE-based optimization strategy. A Multi75E-G probe with a conductive platinum coating was employed.

The VAE models were implemented using the Pyroved Python package. Each model employed a 2-dimensional latent space and fully connected encoder–decoder networks with two hidden layers of 128 units each. Importantly, the dimensionality of the latent space is also a hyperparameter. While a 2D latent space is most convenient for illustrating the workflow due to its ease of visualization, in real studies the dimensionality should also be optimized. Training was performed with mini-batches of size 128 using the Adam optimizer with a learning rate of $10^{-3}$ for 1000 epochs. The input PFM scan was normalized to the range [0, 1] using the 1st and 95th percentiles to reduce the influence of outliers.

GMMs were fitted to the VAE latent distribution using scikit-learn. For each window size and invariance setting, we grid-searched the number of components (K = 2–10) under a full-covariance parameterization. Each candidate was trained with the Expectation–Maximization algorithm (10 random initializations; max_iter = 500), then ranked by log-likelihood, BIC, and ICL; the top-scoring model was retained for analysis.

BGMMs were implemented with the Pyro library. For each window size and invariance setting, we trained an overcomplete mixture with K=20 components and full covariances, using an $LKJ$–$Cholesky$ prior on correlations ($\eta=20$) and per-dimension $HalfNormal(0.05)$ priors on the scales. The component means had $Uniform(-5,5)$ priors and mixture weights had $Dirichlet(\alpha/K)$ with $\alpha = 0.1$. Models were optimized for 500 SVI steps using the $TraceELBO$ with an $AutoDiagonalNormal$ guide and $ClippedAdam$ optimizer (learning rate = $5\times10^{-3}$). A Bhattacharyya-overlap penalty (with $\lambda = 0.1$) was added to discourage component overlap.

**Acknowledgements**

This work was partially supported (BNS, SVK) by the Center for Advanced Materials and Manufacturing (CAMM), the NSF MRSEC center. This work was also partially supported (HF) by the Japan Science and Technology Agency (JST) as part of Adopting Sustainable Partnerships for Innovative Research Ecosystem (ASPIRE) (Grant Number JPMJAP2312), and by MEXT Program: Data Creation and Utilization Type, Material Research and Development Project (Grant Number JPMXP1122683430). The development of the Pyroved Python package (MAZ) was supported by the Laboratory Directed Research and Development Program at Pacific Northwest National Laboratory, a multiprogram national laboratory operated by Battelle for the U.S. Department of Energy.

**Author Contributions**

**BNS**: Conceptualization; Investigation; Software; Writing – original draft. **KB**: Conceptualization; Data curation. **HF**: Resources. **MAZ:** Software; Writing – review & editing. **VVS:** Writing – review & editing. **DCL**: Writing – review & editing. **SVK**: Conceptualization; Supervision; Writing – review & editing.

**Data Availability Statement**

The analysis codes that support the findings of this study are available at https://github.com/Slautin/2025_VAE_star

# Reward driven discovery of the optimal microstructure representations with invariant variational autoencoders

Boris Slautin[1,*], Kamyar Barakati[2], Hiroshi Funakubo[3], Maxim A. Ziatdinov[4], Vladimir V. Shvartsman[1], Doru C. Lupascu[1], Sergei V. Kalinin[2,4*]

[1] Institute for Materials Science and Center for Nanointegration Duisburg-Essen (CENIDE), University of Duisburg-Essen, Essen, 45141, Germany

[2] Department of Materials Science and Engineering, University of Tennessee, Knoxville, TN 37996, USA

[3] Department of Material Science and Engineering, Tokyo Institute of Technology, Yokohama 226-8502, Japan.

[4] Pacific Northwest National Laboratory, Richland, WA 99354, USA

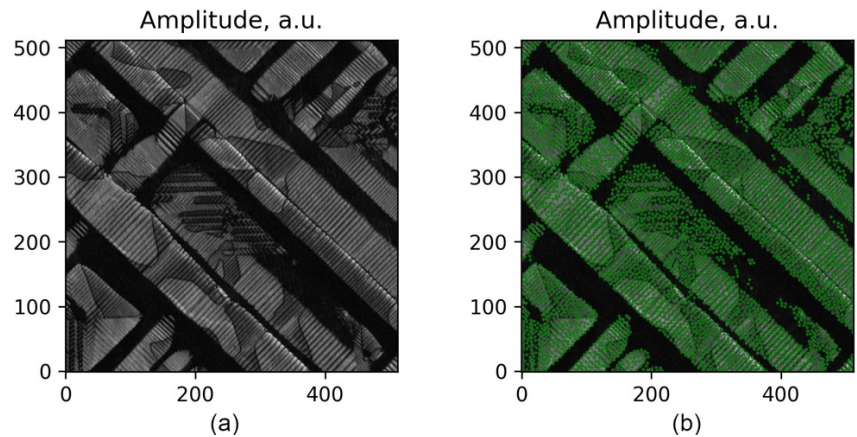

Figure S1: (a) PFM amplitude image with (b) a system of key points selected along the domain walls at a spacing of 5 px.

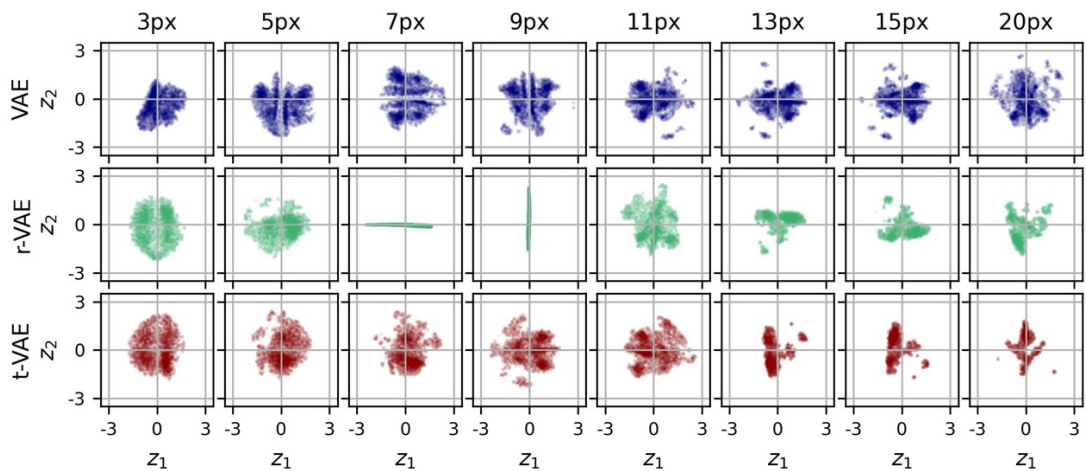

**Figure S2.** Data points filtered from VAE latent space distributions, retaining only those within regions of relative density greater than 0.95 of the maximum. The filtering was performed using kernel density estimation (KDE) with a bandwidth of 0.05.